\documentclass[conference,a4paper]{IEEEtran}
\IEEEoverridecommandlockouts

\usepackage[hidelinks]{hyperref}
\usepackage[cmex10]{amsmath}
\usepackage{amssymb,amsfonts}
\usepackage{dblfloatfix}

\usepackage[ruled,vlined]{algorithm2e}
\usepackage{graphicx}
\graphicspath{{Figures/PDF/}{Figures/PNG/}}

\usepackage{booktabs}
\usepackage{siunitx}
\usepackage[numbers,compress]{natbib}
\usepackage{texnames}
\usepackage{bm,bbm}
\usepackage{orcidlink}
\usepackage{multirow} 
\usepackage{fontawesome5}

\begin{document}

\title{\uppercase{Beluga Whale Detection from Satellite Imagery with Point Labels}
\thanks{This work was supported by the Fourth IEEE GRSS Student Grand Challenge. This work was also supported by the National Earth Observation Data Center Research Projects. Thanks to Prof. A. Camps, Prof. P. Gamba, Dr. D. Kunkee, and Dr. T. Wang for steering the student challenge. Thanks to B. Wu for technical support on ArcGIS Pro. Thanks to Z. Zhang for data acquisition.}
}

\author{	
	\IEEEauthorblockN{Yijie Zheng\orcidlink{0009-0004-5580-0889}\textsuperscript{1,2}, Jinxuan Yang\textsuperscript{3,4}, Yu Chen\textsuperscript{3,5}, Yaxuan Wang\textsuperscript{6,7}, Yihang Lu\textsuperscript{1,2} and Guoqing Li\orcidlink{0000-0003-0654-5426}\textsuperscript{1,\faEnvelope[regular]}}
    
	\IEEEauthorblockA{\textsuperscript{1}\textit{Aerospace Information Research Institute, Chinese Academy of Sciences, Beijing, China}\\
	\textsuperscript{2}\textit{School of Electronic, Electrical and Communication Engineering, University of Chinese Academy of Sciences, Beijing, China}\\
	\textsuperscript{3}\textit{College of Resources and Environment, University of Chinese Academy of Sciences, Beijing, China}\\
	\textsuperscript{4}\textit{Institute of Geographic Sciences and Natural Resources Research, Chinese Academy of Sciences, Beijing, China}\\
	\textsuperscript{5}\textit{Chinese Research Academy of Environmental Sciences, Beijing, China}\\
	\textsuperscript{6}\textit{School of Future Technology, University of Chinese Academy of Sciences, Chinese Academy of Sciences, Beijing, China}\\
	\textsuperscript{7}\textit{Technical Institute of Physics and Chemistry, Chinese Academy of Sciences, Beijing, China}\\
	\textsuperscript{\faEnvelope[regular]}Corresponding author, {ligq@aircas.ac.cn}
	}
}

\maketitle
\begin{abstract}

Very high-resolution (VHR) satellite imagery has emerged as a powerful tool for monitoring marine animals on a large scale. However, existing deep learning-based whale detection methods usually require manually created, high-quality bounding box annotations, which are labor-intensive to produce. Moreover, existing studies often exclude ``uncertain whales'', individuals that have ambiguous appearances in satellite imagery, limiting the applicability of these models in real-world scenarios.
To address these limitations, this study introduces an automated pipeline for detecting beluga whales and harp seals in VHR satellite imagery. The pipeline leverages point annotations and the Segment Anything Model (SAM) to generate precise bounding box annotations, which are used to train YOLOv8 for multiclass detection of certain whales, uncertain whales, and harp seals. Experimental results demonstrated that SAM-generated annotations significantly improved detection performance, achieving higher $\text{F}_\text{1}$-scores compared to traditional buffer-based annotations. YOLOv8 trained on SAM-labeled boxes achieved an overall $\text{F}_\text{1}$-score of 72.2\% for whales overall and 70.3\% for harp seals, with superior performance in dense scenes.
The proposed approach not only reduces the manual effort required for annotation but also enhances the detection of uncertain whales, offering a more comprehensive solution for marine animal monitoring. This method holds great potential for extending to other species, habitats, and remote sensing platforms, as well as for estimating whale biometrics, thereby advancing ecological monitoring and conservation efforts.
The codes for our label and detection pipeline are publicly available at \url{http://github.com/voyagerxvoyagerx/beluga-seeker}.
\end{abstract}


\section{Introduction}
Global climate change, marine pollution, and human activities pose significant threats to marine mammals, including beluga whales (\textit{Delphinapterus leucas}) \cite{watt_population_2021}. As a key species in Arctic and sub-Arctic regions, beluga whale population dynamics serve as indicators of ecological health, making accurate monitoring crucial for conservation, fisheries management, and policy development \cite{belanger_use_2024}.

Traditional whale surveys, such as ship-based \cite{lin_pioneering_2021} or aerial surveys \cite{matthews_estimated_2017} are effective but limited by restricted coverage and potential disturbances to wildlife.  In contrast, very high-resolution (VHR) satellite imagery enables large-scale monitoring with minimal ecological disruption and human involvement \cite{hoschle_potential_2021}.

Satellite imagery has already been used to identify various species of whales \cite{cubaynes_annotating_2023}.  Manual counting has been conducted in regions such as the Eastern High Arctic-Baffin Bay \cite{watt_eastern_2023}, Cumberland Sound \cite{sherbo_using_2024, belanger_use_2024}, and the southern Kara Sea \cite{fretwell_satellite_2023}, providing extensive point annotations. However, manual annotation remains a time-consuming process. 

To accelerate whale counting, a rule-based semi-automated method has been developed to detect beluga whales \cite{iacozza_semi-automated_2024}. While this method reduces manual effort, it often produces multiple segments for individual whales and struggles to differentiate clustered individuals. These limitations emphasize the need for improved detection approaches. 

YOLO (You Only Look Once) \cite{redmon2016you, Jocher_Ultralytics_YOLO_2023}, an object detection algorithm for locating and classifying objects within imagery, has shown promise in detecting whales using satellite \cite{green_gray_2023} and drone imagery \cite{alsaidi_localization_2024}. However, these methods typically require manually created bounding box annotations, a time-intensive process. A common alternative involves generating bounding boxes from point annotations by applying a fixed buffer around each point, such as a 10-meter buffer for gray whales. While this approach simplifies annotation, it is particularly problematic for beluga whales, which often travel in closely packed groups without open water separating individuals. In such cases, fixed buffers will result in bounding boxes encompassing multiple whales, making it challenging to distinguish individual whales in close proximity \cite{green_gray_2023}.

Whale identification in satellite imagery can also be challenging. Whales are only visible at a certain depth (e.g., 2 m in clear water and limited to the surface in turbid water for beluga whales \cite{stewart_estimating_2024}). Limited resolution makes it hard to distinguish whales in high Beaufort sea state conditions. So beluga whale targets in satellite are classified as ``certain'' or ``uncertain'' based on visible features such as body shape, tail flukes, or spatial context \cite{charry_mapping_2021}. Previous studies \cite{green_gray_2023, gaur_whale_2023} have typically excluded uncertain whales from training datasets, which limits the applicability of these models in real-world scenarios.

This study addresses these challenges by (a) leveraging point annotations to automatically generate bounding box annotations and (b) detecting beluga whales and distinguishing them from harp seals (\textit{Pagophilus groenlandicus}) in Arctic waters using VHR satellite imagery. Specifically, we employ the Segment Anything Model (SAM) \cite{kirillov_segment_2023} to generate a precise mask and bounding box for each whale. We then evaluate YOLOv8 \cite{Jocher_Ultralytics_YOLO_2023} for detecting both certain and uncertain whales, as well as harp seals. Additionally, we assess the effectiveness of using bounding boxes generated by SAM, a naive fixed-sized buffer, and SAM-generated boxes that have subsequently been manually refined by experts.

\section{Materials and Methods}
\subsection{Dataset description and preprocessing}
\subsubsection{Imagery description}
Satellite imagery was collected from key summer habitats of beluga whales in Northern Canada (Clearwater Fiord and Churchill River) during 2021–2022. These data were acquired using WorldView-3 and WorldView-2 platforms with resolutions of 0.3 m and 0.46 m, respectively. Details of the imagery and object counts are summarized in Table  \ref{tab:StudyArea}.
\begin{figure}
    \centering
    \includegraphics[width=1\linewidth]{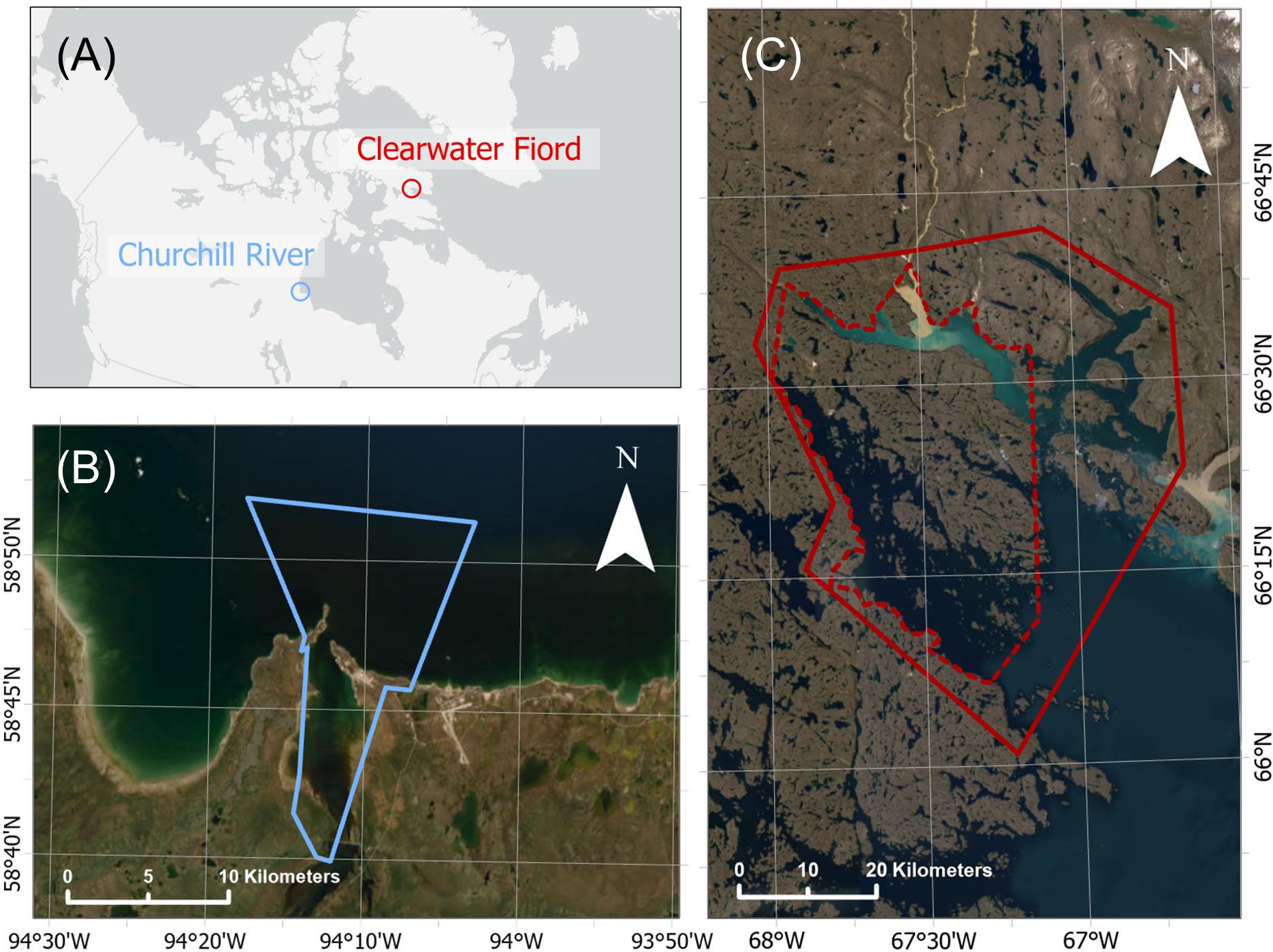}
    \caption{(A) Location of Clearwater Fiord and Churchill River. (B) Blue line outlines the image extent of July 31, 2021 images in Churchill River, covering 146 km² of water. (C) Red line indicates the extent of Aug 1, 2022 satellite images in Clearwater Fiord, with 1281 km$^2$ of water covered. The dashed red line represents the image extent for September 7, 2021, covering 779 km$^2$ of water.}
    \label{fig:enter-label}
\end{figure}

\subsubsection{Point labeling}
The panchromatic satellite imagery was analyzed in ArcMap through a systematic search conducted within a 250 m $\times$ 250 m grid. Objects of interest were categorized as ``certain whale'', ``uncertain whale'', or ``harp seal'' based on the identifiable biometric characteristics (e.g., tail flukes, head, or body shape), or spatial context (e.g., other certain whale targets) \cite{charry_mapping_2021}. Points were placed approximately on the middle of the whale’s body, with close individuals separated using the perpendicular bisector of annotation points. A total of 577 certain whales, 407 uncertain whales, and 542 harp seals were manually annotated.

\begin{table}[t]
    \centering
    \caption{Study Area Details and Number of (short for No.) identified objects from manual counts}\label{tab:StudyArea}
    \begin{tabular}{l c c c}
        \toprule
        Study Area& Clearwater& Clearwater& Churchill \\ \midrule
        Date & 2021 & 2022 & 2022 \\
        Platform & WV-3 & WV-2 & WV-3 \\
        No. certain whales& 196 & 103 & 378 \\
        No. uncertain whales& 215 & 79 & 113 \\
        No. harp seals& 526 & 16 & 0 \\
        \bottomrule
    \end{tabular}
\end{table}

\subsubsection{Image preprocessing}
The panchromatic images, originally with a 16-bit depth, were converted to 8-bit for compatibility with the YOLOv8 model. Images were cropped into smaller patches: 320 × 320 for WV-3 and 192 × 192 for WV-2, ensuring consistent area coverage between these platforms. Only patches containing objects of interest were retained, resulting in a dataset of 538 image patches. 

\subsection{Automated box labeling}
\subsubsection{Segment anything model}
SAM is a general-purpose promptable segmentation model capable of segmenting objects based on user-defined prompts. SAM processes inputs using two main components: an image encoder, which extracts features from the image, and a prompt encoder, which integrates user-provided prompts such as points or bounding boxes. These inputs are then passed to a mask decoder, which generates segmentation masks for the target objects. Trained on a billion-scale dataset, SAM demonstrates strong generalization, including successful applications in remote sensing instance segmentation \cite{wang_samrs_2023}. This study employed SAM-H, the most capable SAM variant.

\subsubsection{Labeling procedure}
\begin{figure}
    \centering
    \includegraphics[width=1\linewidth]{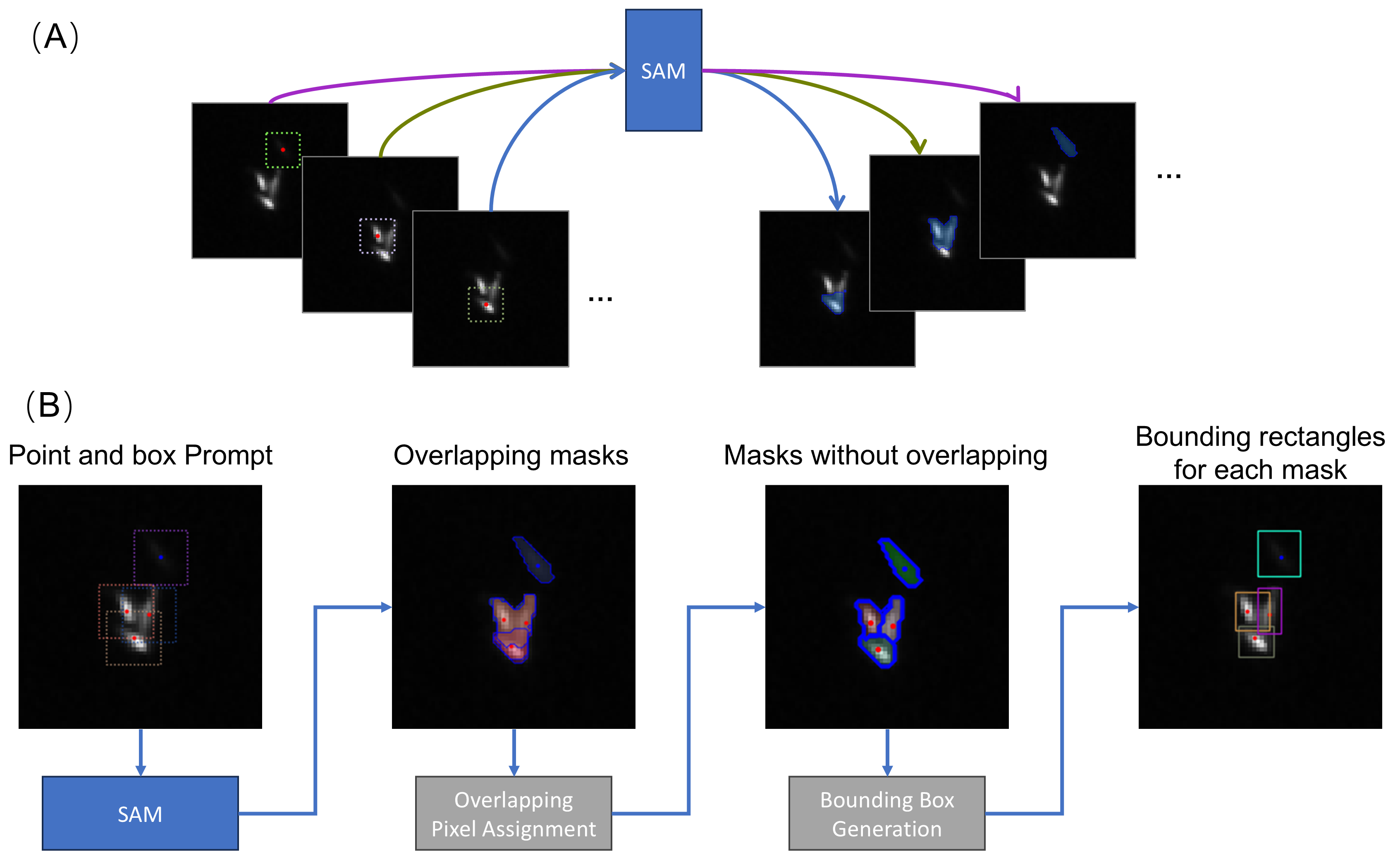}
    \caption{Automated Box Labeling Procedure.}
    \label{fig:label}
\end{figure}

The labeling procedure aimed to generate distinct segmentation masks for each object in the image patches. For each object, the prompt inputs included its point annotation and a buffer box (4 meters for beluga whales and 2 meters for harp seals) around the point. Next, mask for the individual object was generated from the mask decoder (Fig. \ref{fig:label} (A)).

However, SAM could not separate whales that have no open water between them.
For dense scenes where masks overlapped, an overlapping pixel assignment algorithm was applied. Specifically, for each overlapping pixel, the algorithm calculated its Euclidean distance to all annotation points and assigned the pixel to the object whose annotation point was nearest. This ensured that each pixel was uniquely attributed to a single object, resulting in non-overlapping, instance-specific masks. Finally, these masks were converted into bounding rectangles for training object detection models (Fig. \ref{fig:label} (B)). The automatically generated bounding boxes were further refined by experts to assess the quality of the SAM annotations.

\subsection{Beluga whale detection}
\subsubsection{Whale detection model}
YOLO is a single-stage object detection model with proven effectiveness in detecting marine animals in satellite imagery \cite{green_gray_2023,kapoor_deep_2023}. In this study, we utilized YOLOv8s for its balance of speed and accuracy. YOLOv8s predicts a batch of boxes for each input image patch, with a confidence score and category for each box.

\subsubsection{Training scheme}
The dataset of 538 cropped images was split into training, validation, and test sets using a 70:10:20 ratio. Standard data augmentation techniques included in YOLOv8, such as flipping, HSV manipulation, mosaics, and affine transformations, were applied to reduce overfitting. Transfer learning was employed by initializing the YOLOv8 model with a pre-trained weight from the MS-COCO dataset \cite{MSCOCO}. To assess the performance of the proposed SAM-based labeling approach, three YOLOv8 models were trained respectively: one using buffer-based bounding boxes (YOLO-Buffer), one using manually refined SAM-generated bounding boxes (YOLO-Box), and one using raw SAM-labeled bounding boxes (YOLO-SAM). 

\subsubsection{Evaluation protocol}
Model performance was evaluated using precision, recall, $\text{F}_\text{1}$-score, and mean $\text{F}_\text{1}$-score of the three annotation classes. Metrics are also calculated for ``whale overall'' which combined certain and uncertain beluga whales to evaluate detection performance for all whale-like objects. For each model, precisions and recalls were optimized by adjusting the confidence threshold to maximize the $\text{F}_\text{1}$-score of certain whales on the validation set. The IoU threshold for determining true positives (TP) was 0.25. The choice of 0.25 accounted for the sensitivity of small objects, where minor annotation errors would significantly affect the IoU between the annotated box and the predicted box. To ensure statistical robustness, each experiment was conducted five times, with the results reported as the average scores across these runs.

\section{Results and Discussion}
\subsection{Accuracy of the automated box labeling pipeline}
Approximately 19\% of certain whale labels, 21\% of uncertain whale labels, and 4\% of harp seal labels were corrected through manual refinement. Uncertain whale labels exhibited a higher correction rate, with greater deviations observed between the SAM-generated and manually refined bounding boxes. This is primarily due to the challenges SAM faces in accurately segmenting uncertain individuals, which often present weaker visual signals and are more affected by adverse sea state conditions.

\subsection{Beluga whale detection results}
\begin{table*}[!h] 
\centering
\caption{Performance Metrics for Different Annotation Approaches}
\label{tab:metrics}
\begin{tabular}{llllllllllllll}
\hline
\multirow{2}{*}{Annotation} & \multirow{2}{*}{$\text{mF}_\text{1}$} & \multicolumn{3}{c}{Certain   whale}    & \multicolumn{3}{c}{Uncertain whale}     & \multicolumn{3}{c}{Whale overall}      & \multicolumn{3}{c}{Harp seal}           \\ \cline{3-14} 
                            &                      & Precision & Recall & $\text{F}_\text{1}$    & Precision & Recall & $\text{F}_\text{1}$    & Precision & Recall & $\text{F}_\text{1}$    & Precision & Recall & $\text{F}_\text{1}$    \\ \hline
YOLO-Box                 & 60.9     &           70.9 & 59.2 & 64.4 & 63.6 & 43.4 & 51.3 & 84.8 & 66.6 & 74.5 & 62.8 & 72.6 & 67.0  \\
YOLO-Buffer                    & 56.1                 & 69.4      & 47.2   & 56.1  & 49.4      & 51.9   & 50.2  & 78.2      & 63.0   & 69.5  & 55.9      & 71.6   & 61.9  \\
YOLO-SAM                    & 60.4                 & 73.5      & 58.2   & 64.7  & 61.8      & 37.2   & 46.1  & 85.8      & 62.7   & 72.2  & 69.1      & 73.1   & 70.3  \\ \hline
\end{tabular}

\end{table*}

\begin{figure*}[!h]
    \centering
    \includegraphics[width=1\linewidth]{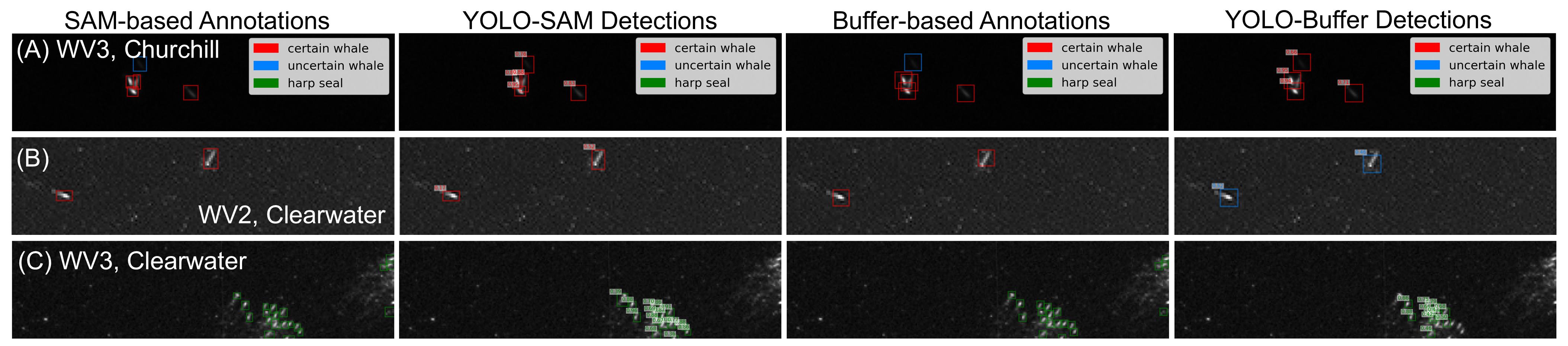}
    \caption{Comparative analysis between SAM-based and buffer box-based annotations (columns 1, 3) and corresponding detection results (columns 2, 4).}
    \label{fig:results}
\end{figure*}

\begin{figure}[!h]
    \centering
    \includegraphics[width=0.55\linewidth]{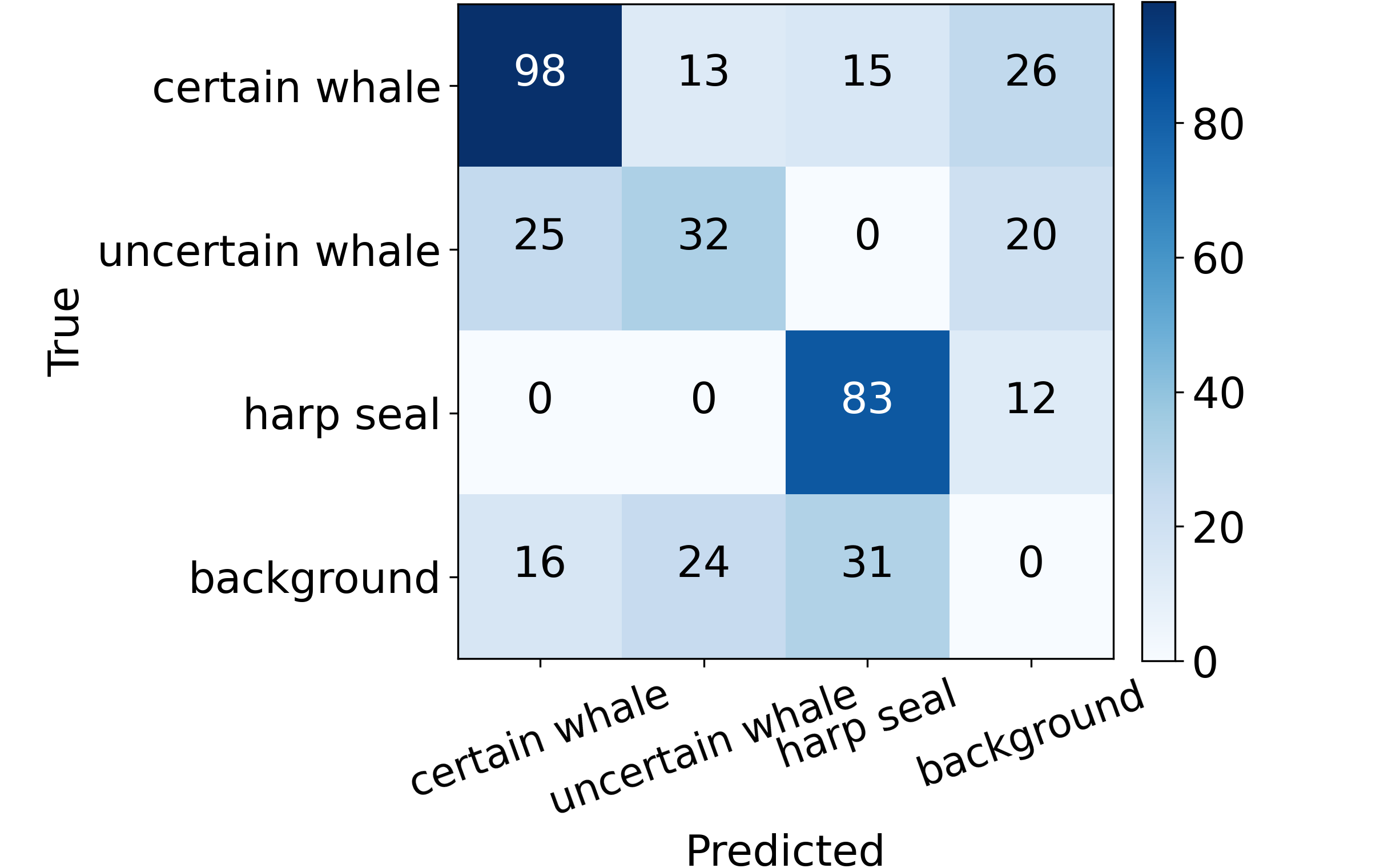}
    \caption{Confusion matrix for beluga whale and harp seal detection.}
    \label{fig:confusion matrix}
\end{figure}

Table \ref{tab:metrics} summarizes the performance metrics of YOLO-Buffer, YOLO-Box and YOLO-SAM on the test set. YOLO-SAM achieved an overall $\text{F}_\text{1}$-score of 72.2\% for whales and 70.3\% for harp seals, demonstrating its effectiveness for detecting these species. Notably, YOLO-SAM significantly outperformed YOLO-Buffer in recall for certain whales (58.2\% vs. 47.2\%) and harp seals (73.1\% vs. 71.6\%), as well as in accuracy across all categories. These results highlighted the advantages of the SAM-based annotation method.

Overall, YOLO-SAM’s mean $\text{F}_\text{1}$-score (m$\text{F}_\text{1}$) closely approached that of YOLO-Box, indicating that the automated labeling pipeline can serve as a viable alternative to manual annotation for detecting beluga whales in satellite imagery.

However, YOLO-SAM yielded lower $\text{F}_\text{1}$-scores for uncertain whales compared to the other models. This reduction in performance was primarily due to the lower annotation quality for uncertain whales. With manual refinement, YOLO-Box achieved $\text{F}_\text{1}$-scores for uncertain whales comparable to those of YOLO-Buffer.

Across all the models, detecting harp seals was easier than identifying certain or uncertain whales, as seals were smaller and typically found in larger groups. In contrast, certain and uncertain whales, being similar in size, posed more challenges for classification due to submersion or occlusion.

Fig. \ref{fig:results} shows qualitative comparisons across platforms and scenes. YOLO-SAM consistently provided more accurate results, with bounding boxes closely matching the whale's body shape. In contrast, YOLO-Buffer produced almost fixed, square-shaped bounding boxes that did not reflect the actual whale size. YOLO-SAM also exhibited fewer missed detections in dense scenes.

Fig. \ref{fig:confusion matrix} presents the confusion matrix for YOLO-SAM's classification results with a confidence threshold of 0.15. The model achieved high precision and recall for certain whales and harp seals but showed misclassifications between whales and the background, as well as between seals and the background. These errors were caused by environmental factors, such as waves, glare, and rocks. Additionally, mild cloud cover could lead to increased missed detections.

\subsection{Discussion}

\subsubsection{Advantages of SAM-labeled bounding boxes}
SAM-labeled bounding boxes offer notable benefits over fixed buffer boxes, contributing to improved detection performance. These bounding boxes closely align with the actual shape of target objects, minimizing the inclusion of irrelevant noise or neighboring objects. This precision enhances the model's focus on learning target-specific features. Additionally, the overlapping pixel assignment algorithm reduces overlap between adjacent boxes, minimizing errors during non-maximum suppression. This results in improved recall of certain whales in crowded scenarios.

\subsubsection{Potential applications}
The SAM-based annotation method can be extended to annotate other species using point labels (e.g., narwhals or polar bears) and applied to aerial remote sensing images. Furthermore, SAM-labeled masks enable the estimation of whale biometrics—including measurements such as body width and length—offering valuable insights for ecological studies \cite{clark_deep_2024}.

\section{Conclusion}
This study proposed an automatic box-labeling pipeline using point labels to detect beluga whales in VHR satellite imagery. By integrating SAM and an overlapping pixel assignment algorithm, this pipeline generates box annotations that closely align with the shape and size of target objects, improving annotation precision while significantly reducing the labor required. When used to train the YOLOv8s model, these annotations led to substantial improvements in detection performance over traditional buffer-based annotations, achieving higher recall and $\text{F}_\text{1}$-scores, particularly for certain whales and harp seals.

Future work could explore extending the pipeline to other species, habitats, and remote sensing platforms to enhance its utility in conservation science. In addition to detecting marine animals, SAM-labeled masks could facilitate the estimation of key biometrics, such as whale width and length. The proposed methods demonstrate strong potential for broader ecological monitoring and conservation applications.

\small
\bibliographystyle{IEEEtranN}
\bibliography{references}

\end{document}